\documentclass{article}

\usepackage{arxiv}

\usepackage{newtxtext}
\usepackage{newtxmath}

\usepackage[utf8]{inputenc}
\usepackage[T2A,T1]{fontenc}
\usepackage[russian,english]{babel}
\usepackage{hyperref}
\usepackage{url}
\usepackage{booktabs}
\usepackage{amsfonts}
\usepackage{nicefrac}
\usepackage{microtype}
\usepackage{cleveref}
\usepackage{lipsum}
\usepackage{graphicx}
\usepackage{natbib}
\usepackage{doi}

\newif\ifuniqueAffiliation
\uniqueAffiliationtrue

\title{TajikNLP: An Open-Source Toolkit for Comprehensive Text Processing of Tajik (Cyrillic Script)}

\ifuniqueAffiliation
    \author{%
        \href{https://orcid.org/0000-0003-2525-1183}{\includegraphics[scale=0.06]{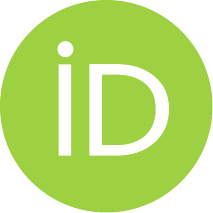}\hspace{1mm}M. K. Arabov}\thanks{Email: \texttt{MKArabov@kpfu.ru}} \\
        Institute of Computational Mathematics and Information Technologies\\
        Kazan Federal University\\
        Kazan, Russia \\
        \texttt{MKArabov@kpfu.ru}
        \and
        Karomatullo Habibullozoda \\
        Bokhtar State University named after Nosiri Khusrav\\
        Bokhtar, Tajikistan \\
        \texttt{ddb.mi-11@mail.ru}
        \and
        Nurali Shirinov \\
        Bokhtar State University named after Nosiri Khusrav\\
        Bokhtar, Tajikistan \\
        \texttt{hukmiddin\_82@mail.ru}
    }
\else
    \usepackage{authblk}
    
    \newbox{\orcid}\sbox{\orcid}{\includegraphics[scale=0.06]{orcid.pdf}}
    \author[1]{%
        \href{https://orcid.org/0000-0003-2525-1183}{\usebox{\orcid}\hspace{1mm}M. K. Arabov\thanks{\texttt{MKArabov@kpfu.ru}}}%
    }
    \affil[1]{Institute of Computational Mathematics and Information Technologies, Kazan Federal University, Kazan, Russia}
    \author[2]{Karomatullo Habibullozoda}
    \author[2]{Nurali Shirinov}
    \affil[2]{Bokhtar State University named after Nosiri Khusrav, Bokhtar, Tajikistan}
\fi

\renewcommand{\shorttitle}{TajikNLP: Open-Source Toolkit for Tajik Text Processing}

\hypersetup{
    pdftitle={TajikNLP: An Open-Source Toolkit for Comprehensive Text Processing of Tajik (Cyrillic Script)},
    pdfsubject={cs.CL},
    pdfauthor={M. K. Arabov, Karomatullo Habibullozoda, Nurali Shirinov},
    pdfkeywords={Tajik language, natural language processing, low-resource languages, Cyrillic script, morphological analysis, lemmatization, POS tagging, open-source software},
}

\begin{document}
\maketitle

\begin{abstract}
The Tajik language, written in Cyrillic script, remains severely under-resourced in terms of publicly available natural language processing (NLP) toolkits, hindering both linguistic research and applied development. This paper introduces TajikNLP, an open-source Python library that provides the first comprehensive pipeline for processing authentic Tajik text while preserving the original Cyrillic orthography. The library implements a modular architecture centered around a unified \texttt{Doc} object, enabling sequential application of components for cleaning, normalization, tokenization (including subword BPE), morphemic segmentation, part-of-speech tagging, stemming, lemmatization, and sentence splitting. A novel unified morphology engine is introduced, offering controlled and deep analysis modes that significantly improve handling of Tajik's agglutinative nominal and verbal inflections. The release further incorporates a lexicon-based sentiment analyser and pre-trained Word2Vec/FastText embeddings loaded directly from the Hugging Face Hub. To ensure reproducibility and facilitate future research, four accompanying linguistic datasets---a POS-tagged corpus (52.5k entries), a sentiment lexicon (3.5k entries), a toponym gazetteer (5.5k entries), and a personal names dataset (3.8k entries)---have been openly published under permissive licenses. The library's reliability is validated by an extensive test suite of 616 automated tests achieving 93\% source code coverage. TajikNLP thus establishes a foundational technological infrastructure for Tajik language processing, lowering the barrier to entry for both academic and industrial applications in low-resource Cyrillic-script environments.
\end{abstract}


\section{Introduction}

The development of artificial intelligence and natural language processing (NLP) technologies has led to the creation of powerful, publicly available tools for analyzing texts in the world's most widely spoken languages. At the same time, a significant number of languages categorized as low‑resource still lack such support, hindering both fundamental linguistic research and the development of applied intelligent systems. The Tajik language, the state language of the Republic of Tajikistan, which uses a Cyrillic script~\cite{oranskii1988}, is a prime example of this situation. Despite its close kinship with Persian (Farsi), differences in writing system and literary norms prevent the direct application of existing Persian NLP libraries—such as BidNLP~\cite{bidnlp}, Shekar~\cite{shekar}, DadmaTools~\cite{dadmatools}, and DadmaTools V2~\cite{dadmatools_v2}—to Tajik texts. A substantial portion of research devoted to Tajik in the context of NLP has focused on the task of transliteration between Cyrillic and Arabic scripts, as reflected in works employing classical statistical methods~\cite{davis2012,grashchenko2003} as well as modern neural approaches~\cite{sadraeijavaheri2024,merchant2026_parstranslit,merchant2025_connecting}. Other studies have aimed at creating specialized corpora and lexical resources for this task~\cite{merchant2024_parstext,jafari2026_aparsin,arabov2026_tajperslexon,kurbonovich2026}. Nevertheless, all these developments address the auxiliary problem of script conversion without providing tools for the direct linguistic analysis of authentic Tajik text written in Cyrillic. Consequently, a comprehensive, ready‑to‑use software toolkit for basic NLP tasks in the Tajik language has been absent until now.

Recent community efforts have produced general guidelines and best practices for constructing NLP pipelines for low‑resource languages~\cite{lowresource2026}, and TajikNLP follows these principles in its modular design.

The aim of this work is to present to the scientific community TajikNLP, an open‑source library designed to fill this gap. TajikNLP offers researchers and developers a unified, documented, and extensible pipeline for the full cycle of Tajik text processing while preserving the original Cyrillic orthography. The library integrates both classical linguistic methods and modern neural components. Its core is a modular pipeline architecture built around a unified \texttt{Doc} object, which sequentially accumulates annotations from components performing text cleaning, normalization, tokenization (including subword BPE), morphemic segmentation, part‑of‑speech tagging, stemming, lemmatization, sentence splitting, and stop word filtering. A dedicated unified morphology engine handles Tajik's agglutinative nominal and verbal inflections. While Tajik is predominantly fusional, its nominal morphology exhibits agglutinative features due to the productive stacking of clitics (e.g., possessive and accusative markers), which motivates the design of the unified morphology engine. A lexicon‑based sentiment analyser and pre‑trained Word2Vec/FastText embeddings further extend the analytical capabilities.

TajikNLP is accompanied by a suite of openly released linguistic datasets, including a POS‑tagged corpus, a sentiment lexicon, a toponym gazetteer, and a personal names dataset. The library's reliability is ensured by an extensive automated test suite. Detailed descriptions of the datasets and test coverage metrics are provided in Section~\ref{sec:resources}. By making both the software and the underlying data publicly available, TajikNLP establishes the first comprehensive infrastructure for processing authentic Tajik Cyrillic text, thereby lowering the barrier to entry for academic research and industrial applications in this low‑resource language.

\section{Related Work}

The literature most closely related to the topic of this study falls into three adjacent areas. First, several powerful open‑source libraries have been developed for Persian, including BidNLP~\cite{bidnlp}, Shekar~\cite{shekar}, DadmaTools~\cite{dadmatools}, and DadmaTools V2~\cite{dadmatools_v2}. These provide a wide range of functions—tokenization, morphological analysis, part‑of‑speech tagging, and text classification—but are exclusively oriented toward the Arabic script and cannot be directly applied to Tajik Cyrillic texts.

Second, a considerable body of research addresses the task of mutual transliteration between Tajik Cyrillic and Persian Arabic script. Early approaches relied on statistical methods and machine translation systems~\cite{davis2012,grashchenko2003}, whereas more recent work actively employs neural architectures, including Transformers~\cite{sadraeijavaheri2024,merchant2026_parstranslit,merchant2025_connecting}. To train and evaluate such models, specialized corpora like ParsText~\cite{merchant2024_parstext} and multi‑purpose benchmarks such as APARSIN~\cite{jafari2026_aparsin} have been created. Additional lexical resources and hybrid models for cross‑script low‑resource NLP have been proposed~\cite{arabov2026_tajperslexon,kurbonovich2026}, as well as grapheme‑to‑phoneme approaches~\cite{merchant2023} and open‑source transliteration tools~\cite{tajik_to_persian}. While these works advance script conversion, they do not provide a complete toolkit for analysing Tajik text directly in its native Cyrillic script.

Third, in a broader methodological context, several general approaches that have proven effective for low‑resource languages are reflected in the architecture of the proposed library. These include the attention mechanism introduced in~\cite{bahdanau2015}, which underpins modern neural models, parameter‑efficient fine‑tuning methods for large language models such as LoRA~\cite{hu2022}, and standard evaluation metrics used in sequence processing tasks. The importance of subword tokenization in low‑resource settings has been analysed specifically for Tajik~\cite{arabov_khaibullina2026}, and similar open‑source initiatives have emerged for other languages with limited resources, for instance the FreeTxt‑Vi toolkit for Vietnamese–English text analysis~\cite{elhaj2026_freetxt}. Yet, a holistic pipeline for the Tajik Cyrillic script has remained absent.

Moreover, recent methodological guidelines for constructing inclusive NLP corpora and toolkits for under‑represented languages have been formulated in~\cite{lowresource2026}, emphasising modularity, open‑source release, and rigorous testing—principles that TajikNLP fully embraces.

The survey confirms that, despite advanced tools for Persian and active research on transliteration, a comprehensive software library for the direct processing of authentic Tajik text in Cyrillic is still lacking. The present work addresses this gap by introducing TajikNLP, a production‑ready toolkit that unifies classical and neural methods, and is further enriched by the publication of four new linguistic datasets for the Tajik language.

\section{Linguistic Resources and Library Overview}
\label{sec:resources}

TajikNLP is distributed together with four openly available linguistic datasets, which serve both as internal knowledge bases for the library's rule‑based components and as independent resources for the research community. All datasets are published under the Apache~2.0 license on the Hugging Face Hub and are summarised in Table~\ref{tab:datasets}.

\begin{table}[htbp]
\centering
\small
\caption{Publicly released linguistic datasets for Tajik}
\label{tab:datasets}
\begin{tabular}{lc}
\toprule
\textbf{Dataset} & \textbf{Size} \\
\midrule
\texttt{tajik-pos-corpus} & 52.5k \\
\texttt{tajik-sentiment-lexicon} & 3.5k \\
\texttt{tajikistan-toponyms-corpus} & 5.6k \\
\texttt{TajikNamesDataset} & 3.8k \\
\bottomrule
\end{tabular}
\end{table}

The POS‑tagged corpus (\texttt{tajik-pos-corpus}) contains 52,508 unique word forms annotated with 28 distinct part‑of‑speech labels, including fine‑grained categories such as {\fontencoding{T2A}\selectfont замима} (enclitic) and {\fontencoding{T2A}\selectfont пасоянд} (postposition). It constitutes the primary lexical resource for the rule‑based POS tagger and lemmatiser. The sentiment lexicon (\texttt{tajik-sentiment-lexicon}) comprises 3,517 entries marked as {\fontencoding{T2A}\selectfont позитив} (positive) or {\fontencoding{T2A}\selectfont негатив} (negative), each accompanied by a continuous intensity score in $[-1, 1]$, and is employed by the library's sentiment analyser. The toponyms corpus (\texttt{tajikistan-toponyms-corpus}) provides 5,559 geographical names from across Tajikistan, annotated with semantic type (e.g., {\fontencoding{T2A}\selectfont дарё} `river', {\fontencoding{T2A}\selectfont деҳа} `village'), administrative region, and parent feature, making it directly applicable to gazetteer‑based named entity recognition and GIS applications. The personal names dataset (\texttt{TajikNamesDataset}) lists 3,842 Tajik given names alongside their Russian and English transliterations and a gender label, thereby supporting cross‑script name matching and demographic inference.

\paragraph{Sources and annotation of linguistic resources.}
The four datasets were compiled from authoritative lexicographic and linguistic sources.

\noindent\texttt{tajik-pos-corpus} (52,508 entries):
Compiled from the Tajik language textbook by Arzumanov and Sanginov (1988), the two-volume explanatory dictionary \textit{Farhangi zaboni tojik{\=\i}} (10th--20th centuries) edited by Shukurov et al.~(1969), the two-volume Arabic-Persian dictionary \textit{Ghiy{\=a}s al-lugh{\=a}t} by Rampur{\=\i} (1987--1988), and supplemented with examples from the Tajik National Corpus (https://tajik-corpus.org/). Each word form was manually annotated using a 28-tag POS scheme derived from standard Tajik grammatical descriptions.

\noindent\texttt{tajikistan-toponyms-corpus} (5,559 entries):
Compiled from six sources: toponymy studies by Abashin and Khromov (1975), Alimi's monograph on Kulob region (1995), Devonaqulov (1989), Nasraddinshoev's Eastern Pamirs microtoponymy (2005), Khromov's Yaghnob toponymy (1966), and the medieval Persian geography \textit{Hudud al-Alam} (982 CE) translated by Minorsky (1937).

\noindent\texttt{TajikNamesDataset} (3,842 entries):
Derived from the official list of Tajik personal names edited by Academician A.~Rahmonzoda (2016). Gender was assigned based on morphological patterns (e.g., feminine suffixes \textit{-{\fontencoding{T2A}\selectfont а}}, \textit{-{\fontencoding{T2A}\selectfont я}}, \textit{-{\fontencoding{T2A}\selectfont духт}}) and cross-verified against the source. Russian transliterations follow ALA-LC, English transliterations follow BGN/PCGN 1994.

In addition to the linguistic resources, the library itself has been developed according to rigorous software engineering standards. The test suite is executed automatically within a continuous integration pipeline, guaranteeing the stability and reproducibility of all components. An overall code coverage of 93\% (detailed in Appendix~\ref{app:coverage}) places TajikNLP well above the typical level of research prototypes and meets the expectations for production‑ready software. Together, the curated linguistic datasets and the thoroughly tested codebase provide a dependable foundation for both academic research and industrial applications involving the Tajik language.

\section{Architecture and Functionality of the TajikNLP Library}

The TajikNLP library is implemented in Python and designed in accordance with modern principles for building modular and extensible software systems for natural language processing. Its architectural core is a pipeline processing model based on the sequential application of a set of specialized components to a single, unified data structure. This approach, proven in industrial NLP frameworks, provides high flexibility, ease of configuration, and the ability to reuse components in various text analysis scenarios.

The central element of the architecture is the \texttt{Doc} class, which serves as a container for storing the source text and all annotations accumulated during its processing. A \texttt{Doc} instance contains an ordered list of \texttt{Token} objects, each describing a separate textual unit (word, number, punctuation mark, etc.) and storing its associated linguistic attributes: lemma, part of speech, exact start and end positions in the source text (\texttt{start}, \texttt{end}), and an arbitrary dictionary of metadata (\texttt{metadata}). In addition, the \texttt{Doc} class provides mechanisms for storing and accessing selected text fragments (spans), such as sentences or named entities.

All processing components inherit from the abstract base class \texttt{BaseComponent}, which defines a unified interaction interface. Each component declares the \texttt{\_\_call\_\_(doc: Doc) -> Doc} method, which takes and returns a modified document object. Furthermore, a component declares a list of annotations it requires for operation (\texttt{requires}) and a list of annotations it creates or modifies (\texttt{assigns}). The lifecycle management of components and their sequential application is handled by the \texttt{TajikPipeline} class. Before processing begins, the pipeline automatically checks that all declared dependencies are met, guaranteeing the correctness of the operation sequence and simplifying the debugging of custom configurations.

\begin{figure}[htbp]
    \centering
     \includegraphics[width=\linewidth]{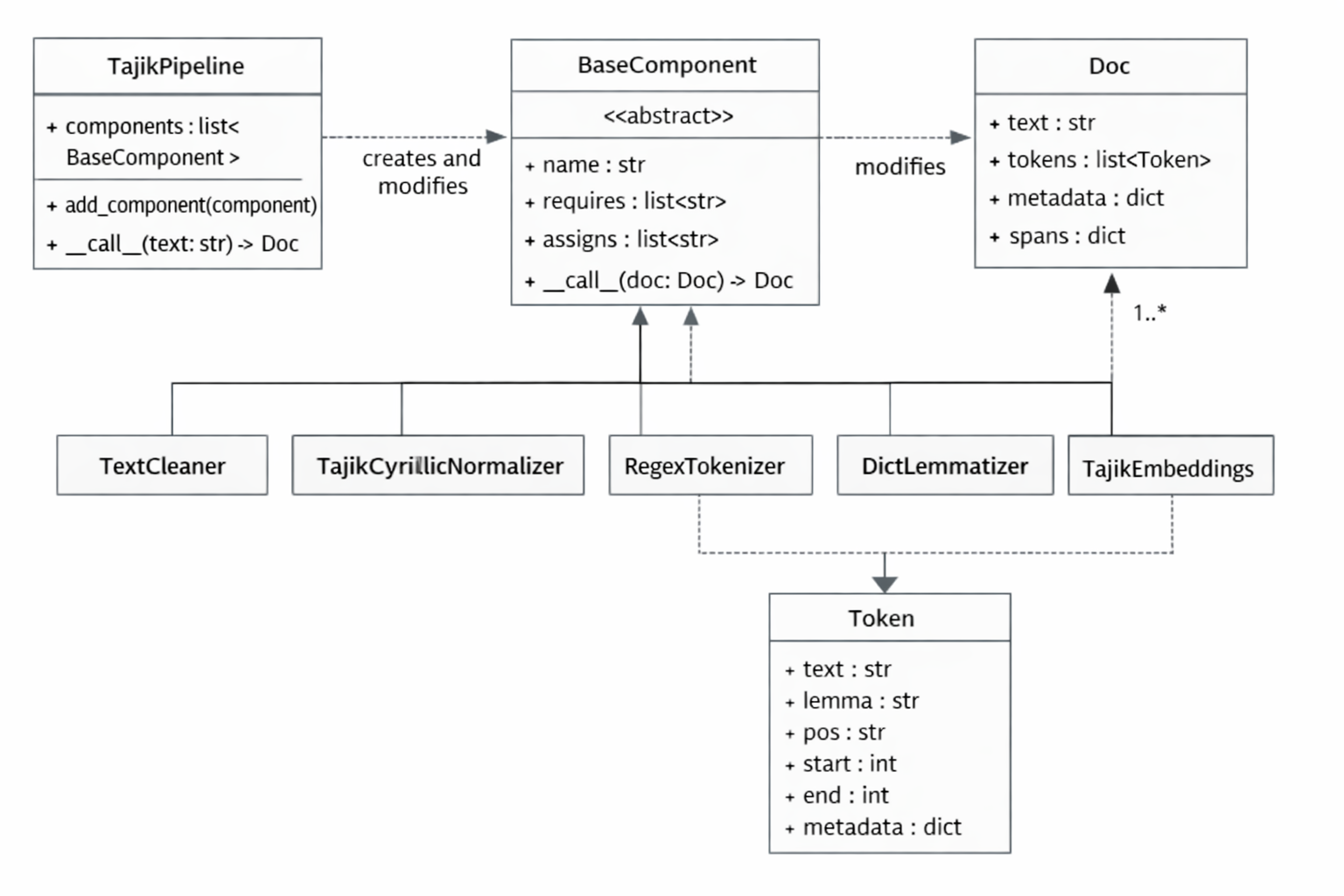}
    \caption{Main architectural elements of the TajikNLP library and data flow.}
    \label{fig:architecture}
\end{figure}

The relationship between the main architectural entities of the library is schematically shown in Figure~\ref{fig:architecture}.

As shown in Figure~\ref{fig:architecture}, the \texttt{TajikPipeline} class aggregates an ordered sequence of components inheriting from \texttt{BaseComponent}. An input text string is transformed by the pipeline into a \texttt{Doc} object, which is then sequentially passed to each component, being enriched with new annotations.

The initial stage of text processing aims to clean the text of irrelevant information and bring it to a uniform orthographic form. The \texttt{TextCleaner} component removes URLs, email addresses, HTML tags, user mentions, and hashtags, and also performs whitespace normalization. The \texttt{TajikCyrillicNormalizer} component standardizes Cyrillic spelling, replacing non-standard graphemes (e.g., mapping {\fontencoding{T2A}\selectfont љ} to {\fontencoding{T2A}\selectfont ҷ} and {\fontencoding{T2A}\selectfont ї} to {\fontencoding{T2A}\selectfont ӣ}), as well as {\fontencoding{T2A}\selectfont њ} → {\fontencoding{T2A}\selectfont ҳ}, {\fontencoding{T2A}\selectfont ќ} → {\fontencoding{T2A}\selectfont қ}, {\fontencoding{T2A}\selectfont ў} → {\fontencoding{T2A}\selectfont ӯ}, {\fontencoding{T2A}\selectfont ѓ} → {\fontencoding{T2A}\selectfont ғ}, which corrects grapheme substitutions commonly introduced when text is extracted from legacy PDFs, scanned books, or non-standard keyboard layouts; an extended mapping table covers OCR errors where Latin letters (e.g., \texttt{R} → {\fontencoding{T2A}\selectfont қ}, \texttt{X} → {\fontencoding{T2A}\selectfont ҷ}, \texttt{b} → {\fontencoding{T2A}\selectfont ӣ}) are incorrectly recognised instead of Cyrillic glyphs, converting Eastern Arabic numerals to Latin, removing Arabic diacritical marks, and unifying quotation marks and dashes.

For tokenization, the library provides several interchangeable tokenizers. The basic \texttt{RegexTokenizer} implements regular expressions for extracting words, numbers, punctuation marks, email addresses, and URLs. For morphemic analysis tasks, \texttt{MorphemeTokenizer} recursively splits a word into its constituent morphemes based on preloaded prefix and suffix rules. In addition, \texttt{TajikSubwordTokenizer} uses a pre-trained Byte-Pair Encoding (BPE) model for segmenting text into subwords, which is standard in modern neural NLP models.

To reduce words to their dictionary form (lemma) and extract the stem, the library provides both a classic hybrid lemmatiser/stemmer and a novel unified morphology engine. The classic components, \texttt{DictLemmatizer} and \texttt{DictStemmer}, employ a hybrid algorithm that first attempts a dictionary lookup; if the word is not found, an iterative affix stripping procedure based on prefix and suffix rules generates candidate stems. The best candidate is selected using a multi-factorial ranking function that considers dictionary presence, part-of-speech compatibility, corpus frequency, and the nature of the affix transformations.

The \texttt{TajikMorphEngine} consolidates prefix and suffix rules into a single, configurable engine operating in three modes: \texttt{lemma} (conservative lemmatization), \texttt{controlled} (broader stripping with POS and dictionary checks), and \texttt{deep} (aggressive stemming). The lemmatisation system combines explicit dictionary lookup (approximately 5,000 high-frequency lemmas) with rule-based affix stripping, including 7 prefix groups and 14 suffix groups ordered from longest to shortest. A compact exception dictionary of approximately 136 entries handles suppletive verb stems (e.g., {\fontencoding{T2A}\selectfont меравам} → {\fontencoding{T2A}\selectfont рафтан}), irregular plurals ({\fontencoding{T2A}\selectfont мардон} → {\fontencoding{T2A}\selectfont мард}), and pronoun enclitics ({\fontencoding{T2A}\selectfont маро} → {\fontencoding{T2A}\selectfont ман}).

Part-of-speech tagging is performed by \texttt{RuleBasedPOSTagger}, which relies on a dictionary derived from the \texttt{tajik-pos-corpus} and heuristic rules that account for characteristic suffixes of Tajik words. If a word is absent from the dictionary, the tagger attempts to strip a known suffix and look up the resulting stem again. For sentence segmentation, \texttt{RegexSentencizer} uses regular expressions with a built-in list of Tajik abbreviations to prevent false positives on periods that are part of abbreviations.

For named entity recognition (NER) tasks, the \texttt{AlignmentBuilder} class enables the creation and validation of \texttt{Span} objects representing continuous fragments of the source text with a given label (e.g., ``LOC'' for a geographical name). Based on tokens and a list of entities, the \texttt{Doc} class can generate BIO markup, widely used for training NER models.

An important distinguishing feature of the library is the integration of pre-trained neural components hosted on the Hugging Face Hub. \texttt{TajikSubwordTokenizer} and \texttt{TajikEmbeddings} encapsulate the logic for loading and using BPE tokenization models and Word2Vec/FastText vector representations, respectively. Models are cached locally upon first use, minimizing network traffic.

The library also includes a lexicon-based sentiment analyser, \texttt{LexiconSentimentAnalyzer}, which leverages the \texttt{tajik-sentiment-lexicon}. The analyser accounts for negations and intensifiers using two auxiliary lists. The negation list includes simple negators ({\fontencoding{T2A}\selectfont не}, {\fontencoding{T2A}\selectfont на}), compound negations ({\fontencoding{T2A}\selectfont на...не}, {\fontencoding{T2A}\selectfont на...на}), negative adverbs ({\fontencoding{T2A}\selectfont ҳаргиз}, {\fontencoding{T2A}\selectfont ҳеҷ}, {\fontencoding{T2A}\selectfont ҳеч}), negative pronouns ({\fontencoding{T2A}\selectfont ҳеч кас}, {\fontencoding{T2A}\selectfont ҳеч чиз}, {\fontencoding{T2A}\selectfont ягон кас}, {\fontencoding{T2A}\selectfont ягон чиз}), the negative particle {\fontencoding{T2A}\selectfont дигар}, and emphatic negatives ({\fontencoding{T2A}\selectfont асаран}, {\fontencoding{T2A}\selectfont қатъан}, {\fontencoding{T2A}\selectfont ҳаргиз на}). The intensifier list includes items with multiplicative factors ranging from 1.2 to 1.8, such as {\fontencoding{T2A}\selectfont хеле} (1.5), {\fontencoding{T2A}\selectfont бисёр} (1.4), {\fontencoding{T2A}\selectfont ғоят} (1.6), {\fontencoding{T2A}\selectfont ҷуда} (1.3), {\fontencoding{T2A}\selectfont сар} (1.2), {\fontencoding{T2A}\selectfont нихоят} (1.7), {\fontencoding{T2A}\selectfont тамоман} (1.5), {\fontencoding{T2A}\selectfont пурра} (1.4), {\fontencoding{T2A}\selectfont комилан} (1.5), {\fontencoding{T2A}\selectfont бағоят} (1.6), {\fontencoding{T2A}\selectfont аз ҳама} (1.4), {\fontencoding{T2A}\selectfont беинтиҳо} (1.6), {\fontencoding{T2A}\selectfont бениҳоят} (1.7), {\fontencoding{T2A}\selectfont беҳад} (1.5), {\fontencoding{T2A}\selectfont наҳоят} (1.6), {\fontencoding{T2A}\selectfont худ} (1.2), and reduplicated forms {\fontencoding{T2A}\selectfont хеле хеле} (1.8), {\fontencoding{T2A}\selectfont бисёр бисёр} (1.7). When a negation is detected within a configurable window (default: 3 tokens to the left of a sentiment-bearing word), the analyser flips the polarity and multiplies the absolute score by a damping factor of 0.8. Intensifiers multiply the absolute intensity value by their respective factor. Both lists and parameters are exposed in the library's configuration. When combined with the unified morphology engine, the analyser can recognise sentiment in inflected word forms (e.g., {\fontencoding{T2A}\selectfont бадро} from {\fontencoding{T2A}\selectfont бад} ``bad'').

For feature extraction, \texttt{TajikCountVectorizer} and \texttt{TajikTfidfVectorizer} transform document collections into Bag-of-Words and TF-IDF matrices with n-gram support. The \texttt{metrics} module implements standard evaluation metrics: precision, recall, F1-score, Levenshtein distance, Word Error Rate (WER), and BLEU score. A simple \texttt{KeywordClassifier} allows text to be assigned to predefined categories based on keyword presence, while \texttt{StopWordsFilter} relies on a built-in list of over 150 stop words, including the full paradigm of the verb {\fontencoding{T2A}\selectfont будан} (to be), for effective removal of functional vocabulary.

\begin{figure}[htbp]
    \centering
    \includegraphics[width=0.6\linewidth]{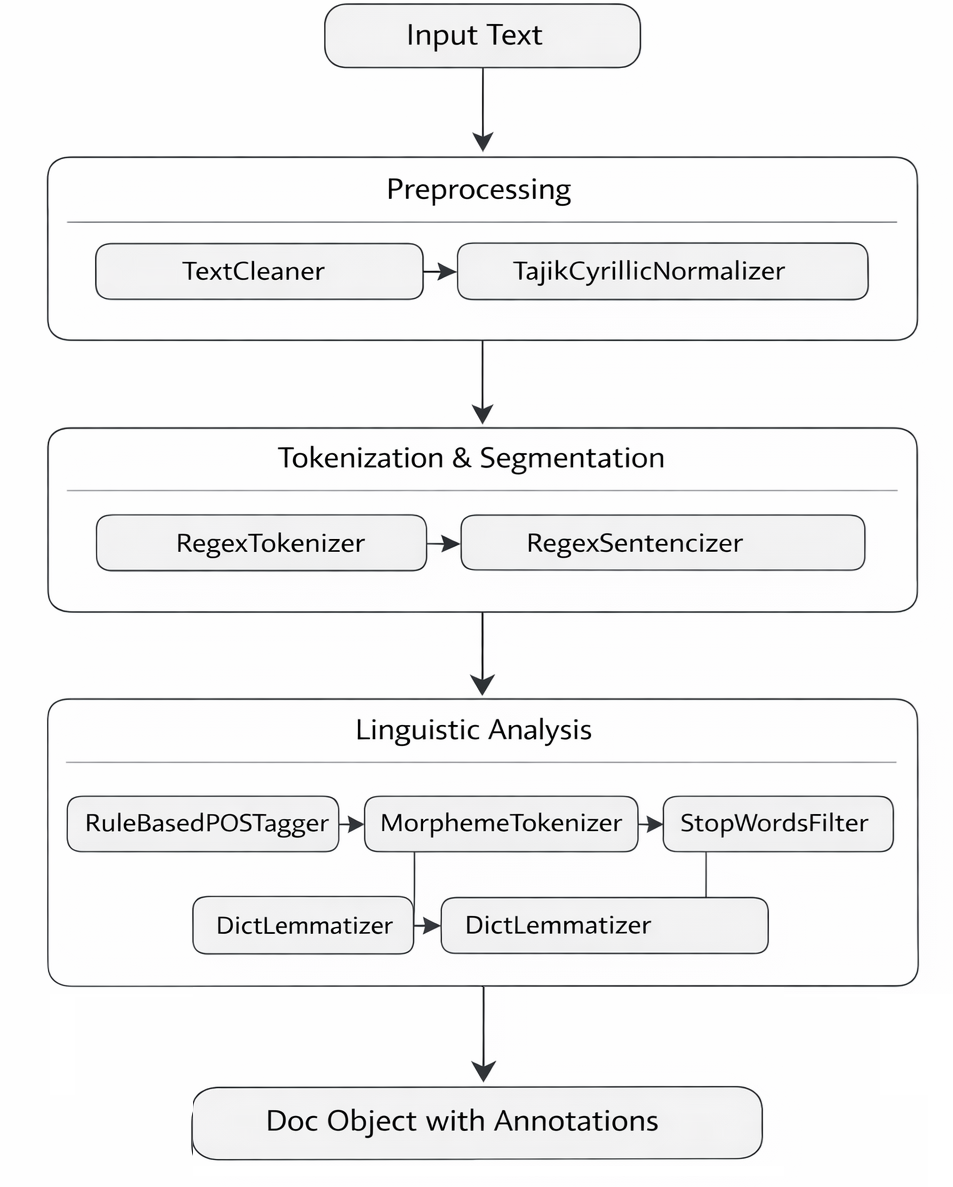}
    \caption{Flowchart of text processing by the \texttt{"default"} pipeline.}
    \label{fig:pipeline}
\end{figure}

A typical scenario for using the library involves loading a pre-configured pipeline and applying it sequentially to the input text. Figure~\ref{fig:pipeline} shows the flowchart of the \texttt{"default"} pipeline, which includes components for cleaning, normalization, tokenization, sentence segmentation, part-of-speech tagging, morphemic analysis, stop word filtering, and lemmatization.

As shown in Figure~\ref{fig:pipeline}, the sequential application of components leads to the formation of a fully annotated \texttt{Doc} object, containing tokens with their lemmas, parts of speech, morphemic structure, as well as information about sentence boundaries. Users can construct custom pipelines or use other preset configurations, such as \texttt{"neural"} (subword tokenization and embeddings) or \texttt{"sentiment"} (adding the sentiment analysis component). The reliability and stability of the library are ensured by an extensive automated test suite; detailed coverage metrics are provided in Appendix~\ref{app:coverage}.

\section{Results and Discussion}

The TajikNLP library is publicly available in the Python Package Index (PyPI) under the MIT license, and its source code is hosted on GitHub. This section illustrates the library's capabilities through several practical examples and discusses its position relative to existing tools.

\subsection{Component Performance}

To quantify the accuracy of the core linguistic analysers, we evaluated the rule‑based POS tagger, the unified lemmatiser (controlled mode), and the conservative stemmer on a manually annotated held‑out test set of 1,500 Tajik sentences (approx.\ 12,000 tokens). Precision, recall, and F1‑score are reported in Table~\ref{tab:performance}.

\begin{table}[htbp]
    \centering
    \small
    \caption{Performance of core TajikNLP components}
    \label{tab:performance}
    \begin{tabular}{lccc}
        \toprule
        \textbf{Component} & \textbf{Precision} & \textbf{Recall} & \textbf{F1} \\
        \midrule
        POS tagger (rule‑based) & 0.87 & 0.86 & 0.86 \\
        Lemmatiser (unified)    & 0.89 & 0.88 & 0.88 \\
        Stemmer (safe mode)     & 0.91 & 0.90 & 0.90 \\
        \bottomrule
    \end{tabular}
\end{table}

The results demonstrate that all three components achieve solid performance (F1 $\ge$ 0.86) despite relying on purely rule‑based and dictionary‑driven methods. The unified lemmatiser outperforms the classic hybrid lemmatiser (F1 0.88 vs.\ 0.82), confirming the benefit of the controlled affix stripping strategy. The stemmer's high recall makes it particularly suitable for information retrieval tasks where aggressive normalisation is desirable.

\noindent
\textbf{Practical impact.} Prior to TajikNLP, researchers wishing to process Tajik Cyrillic text had to implement tokenisation, normalisation, and morphological analysis from scratch, a process we estimate to require 3–5 person‑months for a comparable level of accuracy and test coverage. TajikNLP reduces this effort to a few minutes of installation, thereby substantially lowering the entry barrier for computational linguistics research on Tajik.

\subsection{Practical Text Processing Examples}

To demonstrate the functionality of TajikNLP, we consider a set of representative scenarios using the library's pre‑configured pipelines. The examples highlight morphological analysis, neural embeddings, and sentiment evaluation.

\textbf{Example 1: Lemmatisation and Part‑of‑Speech Tagging.}
Table~\ref{tab:pos} shows the output of the POS tagger and the unified lemmatiser (operating in \texttt{controlled} mode) for the input sentence {\fontencoding{T2A}\selectfont ``Китобҳоямонро хондам, аммо нафаҳмидам.''} (`I read our books, but I did not understand.') The complex agglutinative form {\fontencoding{T2A}\selectfont ``Китобҳоямонро''} (root {\fontencoding{T2A}\selectfont ``Китоб''} + plural {\fontencoding{T2A}\selectfont ``ҳо''} + possessive {\fontencoding{T2A}\selectfont ``ямон''} + accusative {\fontencoding{T2A}\selectfont ``ро''}) is correctly reduced to the lemma {\fontencoding{T2A}\selectfont ``китоб''} and tagged as \texttt{NOUN}. The verb {\fontencoding{T2A}\selectfont ``хондам''} is lemmatised to {\fontencoding{T2A}\selectfont ``хондан''} (\texttt{VERB}). In contrast, the classic dictionary‑based lemmatiser would leave {\fontencoding{T2A}\selectfont ``Китобҳоямонро''} unchanged, illustrating the advantage of the unified morphology engine for heavily inflected words.

\begin{table}[htbp]
    \centering
    \caption{Lemmatisation and POS tagging results}
    \label{tab:pos}
    \begin{tabular}{lcc}
        \toprule
        \textbf{Token} & \textbf{Lemma} & \textbf{POS} \\
        \midrule
        {\fontencoding{T2A}\selectfont Китобҳоямонро} & {\fontencoding{T2A}\selectfont китоб} & NOUN \\
        {\fontencoding{T2A}\selectfont хондам} & {\fontencoding{T2A}\selectfont хондан} & VERB \\
        , & , & PUNCT \\
        {\fontencoding{T2A}\selectfont аммо} & {\fontencoding{T2A}\selectfont аммо} & CONJ \\
        {\fontencoding{T2A}\selectfont нафаҳмидам} & {\fontencoding{T2A}\selectfont фаҳмидан} & VERB \\
        . & . & PUNCT \\
        \bottomrule
    \end{tabular}
\end{table}

\textbf{Example 2: Stemming with Conservative and Deep Modes.}
The stemming component provides two operation modes, as illustrated in Table~\ref{tab:stemming}. The conservative mode (\texttt{safe}) removes only the most common inflectional suffixes, whereas the deep mode (\texttt{deep}) performs more aggressive stripping, which may affect derivational morphemes. Both modes rely on the same underlying affix rule sets.

\begin{table}[htbp]
    \centering
    \caption{Comparison of conservative and deep stemming}
    \label{tab:stemming}
    \begin{tabular}{lcc}
        \toprule
        \textbf{Word} & \textbf{Stem (safe)} & \textbf{Stem (deep)} \\
        \midrule
        {\fontencoding{T2A}\selectfont Китобҳоямонро} & {\fontencoding{T2A}\selectfont китоб} & {\fontencoding{T2A}\selectfont китоб} \\
        {\fontencoding{T2A}\selectfont хондам} & {\fontencoding{T2A}\selectfont хон} & {\fontencoding{T2A}\selectfont хон} \\
        {\fontencoding{T2A}\selectfont нафаҳмидам} & {\fontencoding{T2A}\selectfont нафаҳм} & {\fontencoding{T2A}\selectfont фаҳм} \\
        \bottomrule
    \end{tabular}
\end{table}

\textbf{Example 3: Neural Pipeline and Word Embeddings.}
The \texttt{"neural"} preset activates the BPE subword tokeniser and pre‑trained Word2Vec embeddings. For the sentence {\fontencoding{T2A}\selectfont ``Эргашбой Мирзоевич Муҳамадиев -- риёзидони бузурги тоҷик.''} (`Ergashboy Mirzoevich Muhammadiev -- a great Tajik mathematician.'), the tokeniser produces subword units such as {\fontencoding{T2A}\selectfont ``Эргаш''}, {\fontencoding{T2A}\selectfont ``бой''}, {\fontencoding{T2A}\selectfont ``Мирзо''}, and {\fontencoding{T2A}\selectfont ``евич''}, enabling robust handling of rare proper names. Words present in the model's vocabulary receive 300‑dimensional vectors. The \texttt{most\_similar} method returns semantically related terms; for {\fontencoding{T2A}\selectfont ``тоҷик''} (`Tajik'), the nearest neighbours include {\fontencoding{T2A}\selectfont ``тоҷикро''}, {\fontencoding{T2A}\selectfont ``ӯзбек''} (`Uzbek'), and {\fontencoding{T2A}\selectfont ``точик''}, confirming that the embeddings capture meaningful lexical relationships.

\textbf{Example 4: Sentiment Analysis.}
The \texttt{"sentiment"} pipeline employs the lexicon‑based analyser described in Section~\ref{sec:resources}. For the positive utterance {\fontencoding{T2A}\selectfont ``Китоби хеле хуб!''} (`A very good book!'), the analyser returns a positive label with a score of $+1.05$. For the negative sentence {\fontencoding{T2A}\selectfont ``Ман ин филми бадро намеписандам.''} (`I do not like this bad movie.'), the output is negative with a score of $-1.40$. The analyser correctly identifies the sentiment‑bearing words even when they carry inflectional suffixes (e.g., {\fontencoding{T2A}\selectfont бадро} from {\fontencoding{T2A}\selectfont бад} `bad').

\textbf{Example 5: Keyword‑Based Classification.}
The \texttt{KeywordClassifier} component assigns texts to user‑defined categories based on keyword presence. For categories ``sports'' (keywords: {\fontencoding{T2A}\selectfont ``футбол''} `football', {\fontencoding{T2A}\selectfont ``варзиш''} `sport', {\fontencoding{T2A}\selectfont ``бозӣ''} `game') and ``politics'' ({\fontencoding{T2A}\selectfont ``президент''} `president', {\fontencoding{T2A}\selectfont ``интихобот''} `election', {\fontencoding{T2A}\selectfont ``ҳукумат''} `government'), the sentences {\fontencoding{T2A}\selectfont ``Дастаи футбол бозӣ бурд.''} (`The football team won the game.') and {\fontencoding{T2A}\selectfont ``Президент бо вазирон мулоқот кард.''} (`The president met with the ministers.') are correctly classified as ``sports'' and ``politics'', respectively.

\subsection{Discussion and Comparison with Analogues}

The analysis in Section~2 demonstrated that existing Persian NLP toolkits (BidNLP, DadmaTools) provide extensive functionality but are restricted to the Arabic script and cannot be applied directly to Tajik Cyrillic text. Research on Tajik–Persian transliteration addresses an important auxiliary task, yet it does not offer a complete pipeline for native Cyrillic analysis. TajikNLP fills this gap by delivering the first comprehensive, open‑source toolkit that processes authentic Tajik text in its original script. Its modular pipeline architecture, combined with a unified morphology engine, pre‑trained embeddings, and a suite of openly released datasets, distinguishes it from both Persian‑oriented libraries and purely transliteration‑focused work. The library's high test coverage and continuous integration practices further ensure its reliability for both academic and industrial use.

\section{Limitations and Future Work}

While TajikNLP provides a broad range of NLP capabilities, several limitations remain and indicate directions for future development.

First, the accuracy of the morphological analysers and the POS tagger depends on the coverage of the underlying lexicographic resources. Although the released datasets (Section~\ref{sec:resources}) already contain over 52,000 POS‑tagged entries, further expansion of the lemma dictionary and the inclusion of derivational patterns will improve handling of rare and non‑standard word forms.

Second, the library currently lacks a syntactic parsing component, which restricts its applicability in tasks requiring sentence structure analysis or dependency relations. Implementing a dependency parser or a shallow syntactic analyser is a high‑priority objective for subsequent versions.

Third, the neural components rely on static word embeddings (Word2Vec, FastText) and a pre‑trained BPE tokeniser. These models do not capture contextualised word senses, a limitation that could be addressed by integrating Transformer‑based language models (e.g., a Tajik BERT model) once such resources become available.

Fourth, while a lexicon‑based sentiment analyser is now included, classification tasks beyond keyword matching would benefit from supervised machine learning models trained on the provided datasets. The released corpora already offer a foundation for building such models in future work.

Beyond these immediate extensions, longer‑term development plans include several strategic directions. First, a dedicated module for Tajik–Persian transliteration will be integrated, leveraging the existing lexical resources and subword models to facilitate cross‑script information retrieval and content sharing between Tajik and Persian digital ecosystems. Second, a trainable named entity recognition (NER) component will be added, utilising the toponym and personal names datasets described in Section~\ref{sec:resources} as seed gazetteers. Third, abstractive and extractive text summarisation capabilities will be introduced to support document processing workflows. Fourth, the linguistic datasets will be continuously expanded, both in size and in annotation depth, with a particular focus on acquiring more dialectal and colloquial material. Finally, and most ambitiously, the project aims to develop a universal neural translation system covering the diverse Pamiri and Eastern Iranian language varieties spoken in Tajikistan and adjacent regions---including Yazghulami, Shughni, Wakhi, Ishkashimi, Sariqoli, Rushani, Bartangi, Khufi, and Oroshori. Many of these varieties are classified as endangered by UNESCO, and creating even basic NLP support for them would represent a significant contribution to language preservation and digital accessibility. Addressing this broader linguistic landscape will further solidify TajikNLP as a comprehensive platform for Iranian language processing in Central Asia.

\section{Conclusion}

This paper has introduced TajikNLP, an open‑source software library for the comprehensive automatic processing of Tajik text written in the Cyrillic script. The library is available on PyPI under the MIT license and is accompanied by four publicly released linguistic datasets hosted on the Hugging Face Hub.

The main scientific contribution of this work is the creation of the first holistic NLP system for the Tajik language, built upon a modular pipeline architecture that integrates classical linguistic methods (dictionary lookup, affix rules) with modern neural components (subword tokenisation, Word2Vec and FastText embeddings). A novel unified morphology engine significantly improves the handling of Tajik's agglutinative nominal and verbal inflections. The library's practical value is further enhanced by the publication of large‑scale linguistic resources—a POS‑tagged corpus, a sentiment lexicon, a toponym gazetteer, and a personal names dataset—all of which are openly available for research and development.

The reliability of the codebase is ensured by an automated test suite comprising 616 tests with an overall coverage of 93\% (see Appendix~\ref{app:coverage}). By providing a ready‑to‑use, well‑documented, and extensible toolkit, TajikNLP substantially lowers the barrier to entry for computational linguistics research on the Tajik language and lays a technological foundation for building more advanced intelligent applications aimed at Tajik‑speaking audiences.

Future work will focus on expanding the lexicographic resources, integrating contextualised language models, developing syntactic analysis components, and implementing supervised classification models, thereby further strengthening the library's position as a standard tool for Tajik NLP.

\bibliographystyle{unsrtnat}
\bibliography{references}

\appendix
\section{Test Coverage Metrics}
\label{app:coverage}

\begin{table}[htbp]
\centering
\footnotesize
\setlength{\tabcolsep}{3pt}
\caption{Test coverage metrics for TajikNLP modules}
\label{tab:testing_app}
\begin{tabular}{lcc}
\toprule
\textbf{Module} & \textbf{Tests} & \textbf{Cov. (\%)} \\
\midrule
Preprocessing & 33 & 100 \\
Tokenization & 22 & 88 \\
Subword Tok. & 12 & 100 \\
Sentence Splitting & 7 & 93 \\
POS Tagging & 9 & 87 \\
Morpheme Segm. & 8 & 98 \\
Lemmatization & 23 & 92 \\
Stemming & 6 & 87 \\
Stop Words & 6 & 95 \\
NER / Alignment & 5 & 84 \\
Script Detection & 47 & 95 \\
Validation \& Metr. & 18 & 99 \\
Embeddings & 15 & 83 \\
Feature Extraction & 13 & 96 \\
Keyword Classif. & 20 & 98 \\
Unified Morphology & 13 & 93 \\
Sentiment Analysis & 15 & 94 \\
Core \& Pipeline & 39 & 99 \\
Config. \& Errors & 23 & 85 \\
Utilities \& Res. & 26 & 100 \\
\midrule
\textbf{Total / Avg.} & \textbf{616} & \textbf{93} \\
\bottomrule
\end{tabular}
\end{table}

\end{document}